\documentclass[
]{ceurart}

\usepackage{listings}
\usepackage{todonotes}
\usepackage{multirow}
\usepackage{subfigure}
\usepackage{etoolbox}
\usepackage{array}
\usepackage{tikz}
\usepackage{pgfplots}
\usetikzlibrary{shapes,arrows,arrows.meta, positioning,shapes.misc,decorations.markings,pgfplots.groupplots}
\usepackage{adjustbox}
\usepackage{hyperref}
\usepackage{lipsum,multicol}
\usepackage{xurl}

\usepackage{caption}

\usepackage{mdframed}

\usepackage{etoolbox}
\usepackage[utf8]{inputenc}
\usepackage{pgfplots}
\DeclareUnicodeCharacter{2212}{−}
\usepgfplotslibrary{groupplots,dateplot}
\usetikzlibrary{patterns,shapes.arrows}
\pgfplotsset{compat=newest}

\makeatletter
\patchcmd{\@maketitle}{\raggedright}{\centering}{}{}
\patchcmd{\@maketitle}{\raggedright}{\centering}{}{}
\makeatother

\usepackage{wrapfig}
\usepackage{amsmath}
\usepackage{amsthm}

\usepackage{mathtools}
\usepackage{algorithm}
\usepackage[noend]{algpseudocode}
\usepackage{bbm}

\algnewcommand{\algorithmicgoto}{\textbf{go to}}%
\algnewcommand{\Goto}[1]{\algorithmicgoto~\ref{#1}}%

\begin{document}

\copyrightyear{2023}
\copyrightclause{Copyright for this paper by its authors.
  Use permitted under Creative Commons License Attribution 4.0
  International (CC BY 4.0).}

\conference{CAMLIS'23: Conference on Applied Machine Learning for Information Security,
  October 19--20, 2023, Arlington, VA}

\title{Small Effect Sizes in Malware Detection? Make Harder Train/Test Splits!}

\author[1]{Tirth Patel}[%
email=tpatel9@umbc.edu,
]

\author[1,2]{Fred Lu}[%
email=Lu_Fred@bah.com,
]
\author[1,2]{Edward Raff}[%
email=Raff_Edward@bah.com,
]
\author[1]{Charles Nicholas}[%
email=nicholas@umbc.edu,
]
\author[1]{Cynthia Matuszek}[%
email=cmat@umbc.edu,
]
\author[3]{James Holt}[%
email=holt@lps.umd.edu,
]

\address[1]{University of Maryland, Baltimore County,
  1000 Hilltop Cir, Baltimore, MD 21250}
\address[2]{Booz Allen Hamilton,
8283 Greensboro Drive, McLean, VA 22102}
\address[3]{Laboratory for Physical Sciences, 
5520 Research Park Drive, Catonsville, MD 21228}

\begin{abstract}
Industry practitioners care about small improvements in malware detection accuracy because their models are deployed to hundreds of millions of machines, meaning a 0.1\% change can cause an overwhelming number of false positives. However, academic research is often restrained to public datasets on the order of ten thousand samples and is too small to detect improvements that may be relevant to industry. Working within these constraints, we devise an approach to generate a benchmark of configurable difficulty from a pool of available samples. This is done by leveraging malware family information from tools like AVClass to construct training/test splits that have different generalization rates, as measured by a secondary model. Our experiments will demonstrate that using a less accurate secondary model with disparate features is effective at producing benchmarks for a more sophisticated target model that is under evaluation. We also ablate against alternative designs to show the need for our approach. 
\end{abstract}

\maketitle

\section{Introduction}
\label{sec:introduction}

Malware detection, determining if a given file is benign or malicious, is an important safety problem, since malware causes billions in financial damage each year~\cite{Gantz2014}. However, it is not easy for academic researchers to know that they have produced an improvement using freely available data. This is because industry uses tens of millions of executables at tens of terabytes in scale to detect meaningful improvements in accuracy \cite{Harang2020,Nguyen2021,10.1145/3097983.3098196,Dahl2013a}. In contrast, academic datasets with raw executables available are measured in tens of thousands of executables~\cite{Raff2020a,Eskandari2012,Perdisci2008}. This small scale has made it easy for academic work to over-fit to the data~\cite{raff_ngram_2016,seymor_kaggle_overfit,235493}, and best practices like a train and test set split by time (by when the executable was created) are not possible due to lack of information~\cite{235493}. 

The goal of this work is to provide academic researchers with a means of constructing new train/test splits, using publicly available information for Microsoft windows malware, that can increase the predictive difficulty of the task by removing common biases that lead to overfitting.
The crux of our method is that malware can be grouped into families of related type~\cite{Joyce2022}, and an ideal malware detector is one that can detect new families that were not seen during training. This insight gives us an objective way to group samples into train/test splits that do not cause significant information leakage by having the same malware families in both training and testing, as some prior academic works do~\cite{235493}. By searching for malware families of the right difficulty to place in each train and test split, we can produce new benchmark splits for researchers to use that are smaller than the source datasets, but avoid the bias problems mentioned above. 

The rest of our paper is organized as follows. In \autoref{sec:related_work} we will discuss the important related work to our own, including prior issues in malware detection research and the work in reproducibility and model selection that can be better leveraged by our benchmarks. 
Then in \autoref{sec:method} we will describe how we use a base, simpler model with a search procedure to construct these benchmark datasets. The goal is that our splits will have a lower baseline accuracy for existing methods, showing that we can produce a harder dataset, which in turn makes it easier to detect improvements in generalization and thus effect size. We demonstrate this for three difficulty levels (Easy, Medium, and Hard) in \autoref{sec:results}, and that two intuitive ablation strategies are ineffective in \autoref{sec:ablation}. Finally, our article concludes in \autoref{sec:conclusion}.

\section{Related Work} \label{sec:related_work}

Malware detection research using machine learning has been active since 1995~\cite{Kephart:1995:BID:1625855.1625983}, and includes raw byte~\cite{Kolter:2006:LDC:1248547.1248646}, API calls and assembly~\cite{Shankarapani:2011:MDU:1971266.1971308,Zak2017}, graph~\cite{Kwon:2015:DEI:2810103.2813724}, or exogenous metadata~\cite{Nguyen2019_filename_malicious}. However, much 
industrial research has indicated that academic methods do not often transfer well to industrial data, and so increasingly industry is trying to release more representative datasets~\cite{Anderson2018,Harang2020}. Such efforts are commendable, but these datasets often still require a VirusTotal license\footnote{This costs \$400,000/year.} to get the original files, and they can be prohibitively large. The SOREL-20M corpus has over 20 million files in a train/validation/test split to detect small improvements that matter in real-world use. Our work is the first attempt (that we know of) to develop methods to decrease the amount of data necessary to detect an improvement, rather than simply add more data. 

With respect to the issue of detecting improvements in our models, much of the machine learning literature has tackled this problem. Early works explained that ordinary t-tests and other statistical methods are not reliable for machine learning cases for a variety of technical reasons~\cite{Dietterich:1998:AST:303222.303237}. More recent works have consistently found that a non-parametric Wilcoxon test is a reliable way to detect which algorithm performs best, if multiple trials (i.e., datasets) are available ~\cite{JMLR:v17:benavoli16a,Demsar:2006:SCC:1248547.1248548,dror-etal-2017-replicability}. Other approaches to testing over the space of hyper-parameter values have also been proposed to better measure the improvement achieved, if any, by a new algorithm ~\cite{Bouthillier2021,dror-etal-2019-deep}. The goal of our work is to provide a better foundation for using these prior model selection strategies, as simple cross-validation over an existing biased academic dataset is unlikely to produce a robust conclusion~\cite{VAROQUAUX201868,Varma2006}.

\subsection{Dataset} \label{sec:data}

To perform our study, it was critical that we had a representative population of benign samples, as crawling publicly available sources has been demonstrated to produce models with insufficient diversity, which do not generalize to new malware ~\cite{raff_ngram_2016,seymor_kaggle_overfit,Harang2020,235493}. Because our interest is in producing train/test splits that are also of a reasonable size, so that academics can use them, we use the EMBER 2018 dataset \cite{Anderson2018} which contains 300,000 training and 100,000 testing benign files. The EMBER dataset also includes malicious files, but they are not evenly distributed by malware family or type, which is problematic for our dataset construction approach. 

For this reason, we use the VirusShare corpus~\cite{VirusShare} as a source of freely available malware. Malware family labels can also be obtained freely via the AVClass~\cite{Sebastian2016,Sebastian2020} tool combined with the VirusTotal reports of \cite{Seymour2016}. Using these sources we are able to get hundreds of malware families with thousands of samples each. Following \cite{Raff2020autoyara} we use the same top 184 most frequent malware families with 10,000 samples each, 8,000 for training and 2,000 for testing.

\begin{figure}[!h]
    \centering
    \includegraphics[width=0.8\columnwidth]{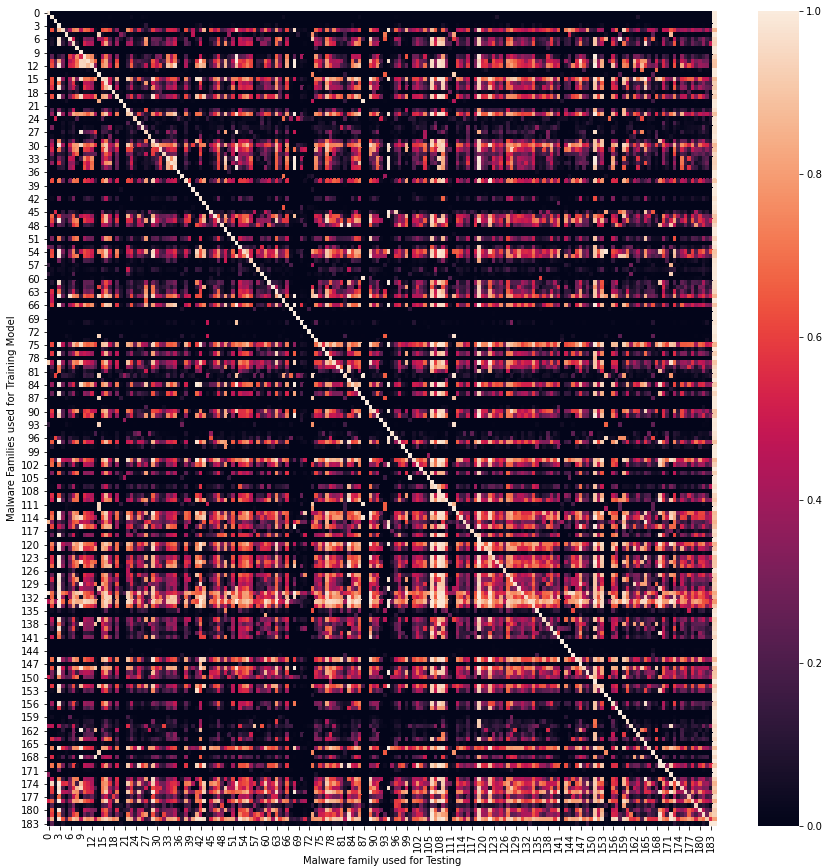}
    \caption{Here we show the cross-error rates of the MalConv models. Each row corresponds to 
    the malware family used in training, and the columns show the recall rate (see color scale) against all the malware families. Dark horizontal bands show malware families that do not generalize to other malware, and vertical dark bands are malware families that are hard to generalize to. Conversely, white bands are easy to generalize from/to respectively. 
    The XGBoost result is near-identical, showing strong correlations in malware family generalization behavior across model and feature types. 
    }
    \label{fig:cross_errors}
\end{figure}

\section{Approach} \label{sec:method}

To make our benchmarks of configurable difficulty, we will start with the all-pairs cross-errors shown in \autoref{fig:cross_errors}. Each row corresponds to selecting one of the 184 malware families to be the only family used during training.  The resulting classifier is then tested on itself and all 183 of the other malware families, with the recall score in the corresponding columns.  (The main diagonal shows the recall we get when testing on the same malware as was used for training). This gives us information on how useful each malware family is, on its own, in predicting all other malware families. In every case a random set of benign files is down-sampled to the same number of malicious files.  That is, 8,000 malicious and 8,000 benign files are used to train a model for each row.

\begin{figure*}
    \centering
    \begin{tikzpicture}

\definecolor{chocolate2267451}{RGB}{226,74,51}
\definecolor{dimgray85}{RGB}{85,85,85}
\definecolor{gainsboro229}{RGB}{229,229,229}
\definecolor{steelblue52138189}{RGB}{52,138,189}

\begin{axis}[
axis background/.style={fill=gainsboro229},
axis line style={white},
tick align=outside,
tick pos=left,
x grid style={white},
xlabel=\textcolor{dimgray85}{Malware Family used for Testing},
xmajorgrids,
xmin=0, xmax=183,
xtick style={color=dimgray85},
xtick={0,1,2,3,4,5,6,7,8,9,10,11,12,13,14,15,16,17,18,19,20,21,22,23,24,25,26,27,28,29,30,31,32,33,34,35,36,37,38,39,40,41,42,43,44,45,46,47,48,49,50,51,52,53,54,55,56,57,58,59,60,61,62,63,64,65,66,67,68,69,70,71,72,73,74,75,76,77,78,79,80,81,82,83,84,85,86,87,88,89,90,91,92,93,94,95,96,97,98,99,100,101,102,103,104,105,106,107,108,109,110,111,112,113,114,115,116,117,118,119,120,121,122,123,124,125,126,127,128,129,130,131,132,133,134,135,136,137,138,139,140,141,142,143,144,145,146,147,148,149,150,151,152,153,154,155,156,157,158,159,160,161,162,163,164,165,166,167,168,169,170,171,172,173,174,175,176,177,178,179,180,181,182,183},
xtick={0,1,2,3,4,5,6,7,8,9,10,11,12,13,14,15,16,17,18,19,20,21,22,23,24,25,26,27,28,29,30,31,32,33,34,35,36,37,38,39,40,41,42,43,44,45,46,47,48,49,50,51,52,53,54,55,56,57,58,59,60,61,62,63,64,65,66,67,68,69,70,71,72,73,74,75,76,77,78,79,80,81,82,83,84,85,86,87,88,89,90,91,92,93,94,95,96,97,98,99,100,101,102,103,104,105,106,107,108,109,110,111,112,113,114,115,116,117,118,119,120,121,122,123,124,125,126,127,128,129,130,131,132,133,134,135,136,137,138,139,140,141,142,143,144,145,146,147,148,149,150,151,152,153,154,155,156,157,158,159,160,161,162,163,164,165,166,167,168,169,170,171,172,173,174,175,176,177,178,179,180,181,182,183},
xticklabel style={rotate=90.0,font=\fontsize{2.5}{4}\selectfont},
xticklabels={FOURshared,adnur,airinstaller,allaple,alman,amonetize,archsms,ardamax,autoit,azero,banbra,bancos,banload,beebone,bettersurf,bifrose,black,blackhole,bladabindi,bredolab,browsefox,bundlore,buterat,buzus,chir,cidox,cinmus,constructor,cosmu,cycbot,dapato,darkkomet,daws,delf,delfinject,delphi,directdownloader,domaiq,dorkbot,downloadadmin,downloadguide,drstwex,dsbot,egroupdial,eorezo,expiro,fakerean,fareit,farfli,fearso,firseria,flystudio,forcestartpage,fosniw,fraudload,fraudpack,fujacks,gabpath,gamarue,gamevance,gepys,geral,hiloti,hlux,hoax,hotbar,hupigon,ibryte,inbox,installbrain,installcore,installerex,installiq,installmonetizer,installmonster,ircbot,karagany,kelihos,klez,kolab,koobface,koutodoor,kraddare,kykymber,ldpinch,lethic,lineage,linkular,linkury,lipler,llac,lmir,loadmoney,lollipop,loring,luder,mabezat,magania,medfos,mediaget,megasearch,menti,monder,morto,mudrop,multiplug,mydoom,nebuler,netsky,nsanti,onlinegames,opencandy,outbrowse,pakes,palevo,parite,pasta,patchload,pcclient,picsys,pincav,poison,powp,prorat,qhost,qqpass,ramnit,rbot,rebhip,refroso,renos,sality,sasfis,scar,sdbot,sefnit,shipup,shiz,sillyfdc,simda,sinowal,skintrim,soft32downloader,softonic,softpulse,somoto,spyeye,staget,swisyn,swizzor,swrort,sytro,tdss,tibs,toggle,trymedia,turkojan,unruy,uptodown,urelas,usteal,vapsup,vbinder,viking,vilsel,virlock,virut,vittalia,vobfus,vtflooder,vundo,wapomi,webprefix,winwebsec,xpaj,xtrat,yakes,zapchast,zbot,zegost,zeroaccess,zlob,zusy,zwangi},
xticklabels={FOURshared,adnur,airinstaller,allaple,alman,amonetize,archsms,ardamax,autoit,azero,banbra,bancos,banload,beebone,bettersurf,bifrose,black,blackhole,bladabindi,bredolab,browsefox,bundlore,buterat,buzus,chir,cidox,cinmus,constructor,cosmu,cycbot,dapato,darkkomet,daws,delf,delfinject,delphi,directdownloader,domaiq,dorkbot,downloadadmin,downloadguide,drstwex,dsbot,egroupdial,eorezo,expiro,fakerean,fareit,farfli,fearso,firseria,flystudio,forcestartpage,fosniw,fraudload,fraudpack,fujacks,gabpath,gamarue,gamevance,gepys,geral,hiloti,hlux,hoax,hotbar,hupigon,ibryte,inbox,installbrain,installcore,installerex,installiq,installmonetizer,installmonster,ircbot,karagany,kelihos,klez,kolab,koobface,koutodoor,kraddare,kykymber,ldpinch,lethic,lineage,linkular,linkury,lipler,llac,lmir,loadmoney,lollipop,loring,luder,mabezat,magania,medfos,mediaget,megasearch,menti,monder,morto,mudrop,multiplug,mydoom,nebuler,netsky,nsanti,onlinegames,opencandy,outbrowse,pakes,palevo,parite,pasta,patchload,pcclient,picsys,pincav,poison,powp,prorat,qhost,qqpass,ramnit,rbot,rebhip,refroso,renos,sality,sasfis,scar,sdbot,sefnit,shipup,shiz,sillyfdc,simda,sinowal,skintrim,soft32downloader,softonic,softpulse,somoto,spyeye,staget,swisyn,swizzor,swrort,sytro,tdss,tibs,toggle,trymedia,turkojan,unruy,uptodown,urelas,usteal,vapsup,vbinder,viking,vilsel,virlock,virut,vittalia,vobfus,vtflooder,vundo,wapomi,webprefix,winwebsec,xpaj,xtrat,yakes,zapchast,zbot,zegost,zeroaccess,zlob,zusy,zwangi},
y grid style={white},
ylabel=\textcolor{dimgray85}{Recall value},
ymajorgrids,
ymin=-0.048425, ymax=1.049925,
ytick style={color=dimgray85},
height=0.4\columnwidth,
width=\textwidth
]
\addplot [line width=0.44pt, chocolate2267451, mark=*, mark size=1, mark options={solid}]
table {%
0 0.338
1 0.744
2 0.28
3 0.978
4 0.697
5 0.5075
6 0.5745
7 0.843
8 0.8585
9 0.998
10 0.7515
11 0.711
12 0.851
13 0.8805
14 0.3545
15 0.927
16 0.845
17 0.8225
18 0.328
19 0.8775
20 0.038
21 0.1955
22 0.9805
23 0.912
24 0.5455
25 0.9515
26 0.6855
27 0.443
28 0.386
29 0.9815
30 0.779
31 0.7005
32 0.25
33 0.8105
34 0.9095
35 0.894
36 0.1995
37 0.2085
38 0.932
39 0.236
40 0.608
41 0.9545
42 0.9765
43 0.2825
44 0.2555
45 0.6825
46 0.792
47 0.8775
48 0.6255
49 0.9245
50 0.6845
51 0.7995
52 0.9975
53 0.8085
54 0.8675
55 0.924
56 0.763
57 0.589
58 0.6995
59 0.678
60 0.7855
61 0.861
62 0.823
63 0.8715
64 0.491
65 0.693
66 0.8655
67 0.5295
68 0.933
69 0.424
70 0.458
71 0.3075
72 0.1665
73 0.3735
74 0.4685
75 0.9475
76 0.8735
77 0.831
78 0.9055
79 0.9205
80 0.8825
81 0.874
82 0.472
83 0.944
84 0.764
85 0.9775
86 0.915
87 0.0025
88 0.0015
89 0.9185
90 0.844
91 0.782
92 0.7415
93 0.349
94 1
95 0.3955
96 0.817
97 0.7405
98 0.5495
99 0.26
100 0.1175
101 0.9615
102 0.8075
103 0.742
104 0.851
105 0.3665
106 0.99
107 0.869
108 0.9835
109 0.9775
110 0.8085
111 0.0195
112 0.1605
113 0.8695
114 0.884
115 0.7095
116 0.7735
117 0.2475
118 0.7705
119 1
120 0.846
121 0.7465
122 0.9965
123 0.681
124 0.718
125 0.8275
126 0.489
127 0.9255
128 0.909
129 0.9275
130 0.939
131 0.9155
132 0.8735
133 0.938
134 0.9285
135 0.545
136 0.817
137 0.8725
138 0.917
139 0.8005
140 0.8135
141 0.877
142 0.0575
143 0.5265
144 0.431
145 0.6915
146 0.9035
147 0.9855
148 0.785
149 0.809
150 0.7475
151 0.996
152 0.884
153 0.978
154 0.113
155 0.6075
156 0.929
157 0.9355
158 0.342
159 0.9615
160 0.877
161 0.4145
162 0.858
163 0.7125
164 0.9395
165 0.863
166 0.8515
167 0.449
168 0.8635
169 0.934
170 0.871
171 0.686
172 0.9045
173 0.9255
174 0.6105
175 0.92
176 0.7725
177 0.828
178 0.7935
179 0.654
180 0.8305
181 0.7155
182 0.8835
183 0.512
};
\addplot [line width=0.44pt, steelblue52138189, mark=*, mark size=1, mark options={solid}]
table {%
0 0.338
1 0.744
2 0.28
3 0.978
4 0.697
5 0.5075
6 0.5745
7 0.843
8 0.8585
9 0.998
10 0.7515
11 0.711
12 0.851
13 0.8805
14 0.3545
15 0.927
16 0.845
17 0.8225
18 0.328
19 0.8775
20 0.038
21 0.1955
22 0.9805
23 0.912
24 0.5455
25 0.9515
26 0.6855
27 0.443
28 0.386
29 0.9815
30 0.779
31 0.7005
32 0.25
33 0.8105
34 0.9095
35 0.894
36 0.1995
37 0.2085
38 0.932
39 0.236
40 0.608
41 0.9545
42 0.9765
43 0.2825
44 0.2555
45 0.6825
46 0.792
47 0.8775
48 0.6255
49 0.9245
50 0.6845
51 0.7995
52 0.9975
53 0.8085
54 0.8675
55 0.924
56 0.763
57 0.589
58 0.6995
59 0.678
60 0.7855
61 0.861
62 0.823
63 0.8715
64 0.491
65 0.693
66 0.8655
67 0.5295
68 0.933
69 0.424
70 0.458
71 0.3075
72 0.1665
73 0.3735
74 0.4685
75 0.9475
76 0.8735
77 0.831
78 0.9055
79 0.9205
80 0.8825
81 0.874
82 0.472
83 0.944
84 0.764
85 0.9775
86 0.915
87 0.0025
88 0.0015
89 0.9185
90 0.844
91 0.782
92 0.7415
93 0.349
94 1
95 0.3955
96 0.817
97 0.7405
98 0.5495
99 0.26
100 0.1175
101 0.9615
102 0.8075
103 0.742
104 0.851
105 0.3665
106 0.99
107 0.869
108 0.9835
109 0.9775
110 0.8085
111 0.0195
112 0.1605
113 0.8695
114 0.884
115 0.7095
116 0.7735
117 0.2475
118 0.7705
119 1
120 0.846
121 0.7465
122 0.9965
123 0.681
124 0.718
125 0.8275
126 0.489
127 0.9255
128 0.909
129 0.9275
130 0.939
131 0.9155
132 0.8735
133 0.938
134 0.9285
135 0.545
136 0.817
137 0.8725
138 0.917
139 0.8005
140 0.8135
141 0.877
142 0.0575
143 0.5265
144 0.431
145 0.6915
146 0.9035
147 0.9855
148 0.785
149 0.809
150 0.7475
151 0.996
152 0.884
153 0.978
154 0.113
155 0.6075
156 0.929
157 0.9355
158 0.342
159 0.9615
160 0.877
161 0.4145
162 0.858
163 0.7125
164 0.9395
165 0.863
166 0.8515
167 0.449
168 0.8635
169 0.934
170 0.871
171 0.686
172 0.9045
173 0.9255
174 0.6105
175 0.92
176 0.7725
177 0.828
178 0.7935
179 0.654
180 0.8305
181 0.7155
182 0.8835
183 0.512
};
\end{axis}

\end{tikzpicture}
    \caption{Example showing the malware recall rate (y-axis) of a model trained using the top five best malware families (in terms of highest average recall against other families) as reported by \autoref{fig:cross_errors}. While the average recall rate is reasonably large, the recall per-family has an extremely high variance. This makes it challenging to determine if a performance difference comes from luck or true effect.}
    \label{fig:best5}
\end{figure*}
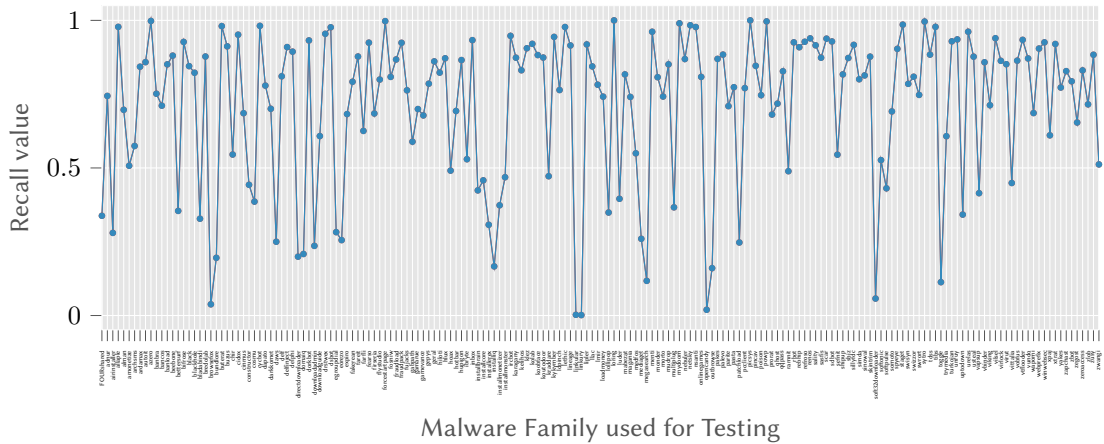

The goal will be to generate train/test splits into three categories mentioned earlier, namely Easy, Medium, and Hard. We will start by generating ten different train/test splits in each of the categories. Note here that each train/test split must have no overlap of malware families between the train and test splits, but different train/test splits might share some families. That is to say, the family ``cycbot'' may occur in training splits 1, 3, and 4, but that means the ``cycbot'' family cannot occur in test splits 1, 3, and 4. In this way every individual split is a meaningful test of generalization to new malware families, the ultimate goal of any malware detector. Each training split is trained on independently (not cross-validated), and so overlap between splits will not impact the results.  
So we will have 30 distinct train/test splits in total, ten each for the Easy, Medium, and Hard categories.

\begin{algorithm}[!h]
\caption{Benchmark search}
\label{algo1}
\begin{algorithmic}[1]
\Require $184\times 184$ accuracy matrix $M$, target recall threshold $\tau$, closeness parameter $\epsilon$, max iterations $I$
\State $T, V \gets \{\cdot \}, \{\cdot \}$ \Comment{Training and validation sets}
\State $C = \{(t_1, v_1), (t_2, v_2), \ldots\} \gets \mathrm{argwhere}(|M - \tau| \leq \epsilon)$ \label{marker}
\State $i = 0$ 
\For{$i \in [ 1, \ldots, 10]$}
    \State Select a new $(t_i, v_i)$ from $C$
    \If{$t_i \in T$ or $v_i \in V$}
        \State Discard $(t_i, v_i)$
    \EndIf
    \If{$|M[t_j, v_i] - \tau| > \epsilon$ for any $t_j \in T$}
        \State Discard $(t_i, v_i)$
    \EndIf
    \If{$|M[t_i, v_j] - \tau| > \epsilon$ for any $v_j \in V$}
        \State Discard $(t_i, v_i)$
    \EndIf
    \If{$(t_i, v_i)$ not discarded}
        \State Add($T, t_i$), Add($V, v_i$)
    \EndIf
    \If{$i > I$}
        \State $\epsilon = \epsilon + 0.05$, then \Goto{marker}
    \EndIf
\EndFor
\State \Return $T, V$
\end{algorithmic}
\end{algorithm}

The algorithm we use to create these train/test splits is shown in \autoref{algo1}. The strategy is to apply a random search to obtain a set of training families $T$ and a set of testing families $V$, which satisfy the constraint that none of the training families perform much better or worse than target recall threshold $\tau$ on any of the testing families. That is, $|M[t_i, v_j] - \tau| < \epsilon$ for $t_i\in T$, $v_j\in V$. This is done from the Malconv 184 x 184 data $M$ by first identifying elements in the matrix which are $\epsilon$-close to $\tau$. The candidate pairs $(t_i, v_i)$ of training and testing families corresponding to those elements are then randomly sampled. 

At each iteration, while the sampled pair satisfies the performance constraint by design, it must also satisfy the constraint pairwise among all the families already selected in the training and testing sets. If this condition holds and neither member of the pair is already selected, then the pair is added to the growing training and testing sets. This procedure runs until 10 distinct families have been chosen for both training and testing. If the algorithm is unable to converge for $\epsilon$, then $\epsilon$ is loosened (increased) and we try again. For efficiency, when this happens we do not discard the progress we have already made, and in line~\ref{marker} we use the previous $\epsilon$ as a lower bound and the new $\epsilon$ as an upper bound for identifying new candidate pairs.
For the Easy, Medium, and Hard splits we use $\tau = $ 0.9, 0.5, and 0.25 respectively. We set the number of iterations $I=1000$ and $\epsilon = 0.05$ throughout. 

\section{Results} \label{sec:results}

In our tests, we use four models for evaluation. First is a byte 6-gram model that has been popular in academic malware detection research for several years~\cite{Kolter:2006:LDC:1248547.1248646,raff_ngram_2016,Kilograms_2019}. Second we use MalConv~\cite{MalConv} and its extended approach MalConvGCT~\cite{Raff2020b}. Finally, we use the Ember feature vectors~\cite{Anderson2018} with the XGBoost algorithm~\cite{xgboost} as the standard domain knowledge approach, which we will refer to as just ``XGBoost'' for brevity. In our experiments, each train/test split we produce has 160,000 and 40,000 total samples, respectively. However, because within a given train/test split no family is used for both training and testing, we note that if memory is a constraint, the experiments can be performed using just the training or testing sets alone. We remind the reader that randomly sampling to a split of this size will produce a model with $\geq 90\%$ accuracy in all cases. 

Having defined our approach to generating harder train/test splits, we begin with the primary results as shown in \autoref{tbl:result}. As can be seen, we are able to successfully produce datasets that are more challenging than the original dataset. With a lower baseline level of accuracy, it becomes possible to measure effect sizes with a moderate number of samples - and avoid the over 30 million files that are needed to reliably detect improvement of XGBoost like models on regular malware data \cite{Harang2020}. 

\begin{table}[!ht]
\caption{A set of Easy, Medium, and Hard train/test splits (``Modified'' columns) created using \autoref{algo1}. The goal is to produce splits that have lower accuracy than the normal benign/malicious classification task (``Normal'' column). In each case, we see that we successfully produce harder splits, which may allow detection of larger effects.}
\label{tbl:result}
\centering
\begin{tabular}{@{}lcccc@{}}
\toprule
\multicolumn{1}{c}{}          &        & \multicolumn{3}{c}{Modified Train/Test} \\ \cmidrule(l){3-5} 
\multicolumn{1}{c}{Algorithm} & Normal & Easy        & Medium       & Hard       \\ \midrule
Byte n-grams                  & 94.87  & 79.48       & 66.06        & 58.52      \\
MalConv                       & 91.14  & 85.88       & 63.81        & 44.73      \\
MalConv GCT                   & 93.29  & 83.43       & 61.51        & 33.49      \\
XGBoost      & 99.64  & 99.08       & 90.80        & 72.80      \\ \bottomrule
\end{tabular}
\end{table}

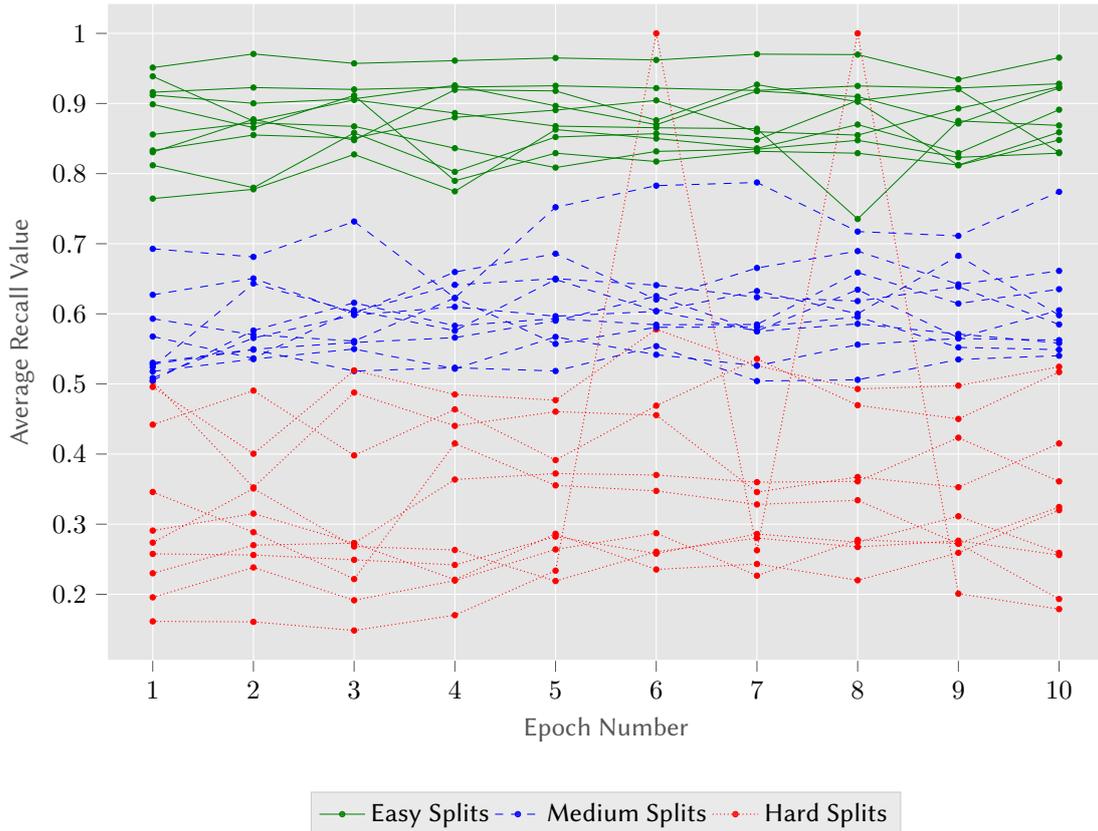
\begin{figure}[!ht]
\begin{tikzpicture}

\definecolor{dimgray85}{RGB}{85,85,85}
\definecolor{gainsboro229}{RGB}{229,229,229}
\definecolor{green}{RGB}{0,128,0}
\definecolor{lightgray204}{RGB}{204,204,204}

\begin{axis}[
axis background/.style={fill=gainsboro229},
axis line style={white},
legend cell align={left},
legend style={
  fill opacity=0.8,
  draw opacity=1,
  text opacity=1,
  at={(0.5,-0.2)},
  anchor=north,
  legend columns=3, 
  draw=lightgray204,
  fill=gainsboro229
},
tick align=outside,
tick pos=left,
x grid style={white},
xlabel=\textcolor{dimgray85}{Epoch Number},
xmajorgrids,
xmin=0.55, xmax=10.45,
xtick style={color=dimgray85},
y grid style={white},
ylabel=\textcolor{dimgray85}{Average Recall Value},
ymajorgrids,
ymin=0.10582, ymax=1.04258,
ytick style={color=dimgray85},
height=0.7\columnwidth,
width=\columnwidth,
]
\addplot [very thin, green, mark=*, mark size=1, mark options={solid}]
table {%
1 0.9387
2 0.8748
3 0.90515
4 0.88635
5 0.86775
6 0.8654
7 0.864
8 0.7352
9 0.8752
10 0.8687
};
\addlegendentry{Easy Splits}
\addplot [very thin, blue, dashed, mark=*, mark size=1, mark options={solid}]
table {%
1 0.52815
2 0.54885
3 0.51835
4 0.52315
5 0.51845
6 0.55395
7 0.5042
8 0.5059
9 0.53495
10 0.54025
};
\addlegendentry{Medium Splits}
\addplot [ red,  densely dotted, mark=*, mark size=1, mark options={solid}]
table {%
1 0.50385
2 0.35235
3 0.4877
4 0.44025
5 0.4605
6 0.4555
7 0.3458
8 0.3673
9 0.3527
10 0.4151
};
\addlegendentry{Hard Splits}
\addplot [very thin, green, mark=*, mark size=1, mark options={solid}]
table {%
1 0.8558
2 0.87245
3 0.86735
4 0.8362
5 0.8086
6 0.8317
7 0.8347
8 0.8475
9 0.8232
10 0.82915
};
\addplot [very thin, blue, dashed, mark=*, mark size=1, mark options={solid}]
table {%
1 0.6927
2 0.6812
3 0.7317
4 0.6226
5 0.75205
6 0.7828
7 0.78735
8 0.71735
9 0.7112
10 0.774
};
\addplot [ red,  densely dotted, mark=*, mark size=1, mark options={solid}]
table {%
1 0.23
2 0.27005
3 0.27305
4 0.2208
5 0.2863
6 0.2354
7 0.24325
8 0.22
9 0.2592
10 0.32
};
\addplot [very thin, green, mark=*, mark size=1, mark options={solid}]
table {%
1 0.9161
2 0.92275
3 0.92005
4 0.9234
5 0.9253
6 0.92195
7 0.9186
8 0.92505
9 0.92205
10 0.92815
};
\addplot [very thin, blue, dashed, mark=*, mark size=1, mark options={solid}]
table {%
1 0.50855
2 0.56555
3 0.59825
4 0.6099
5 0.59665
6 0.60355
7 0.6325
8 0.6001
9 0.68275
10 0.59805
};
\addplot [ red,  densely dotted, mark=*, mark size=1, mark options={solid}]
table {%
1 0.4958
2 0.4005
3 0.5193
4 0.485
5 0.47685
6 0.5779
7 0.52625
8 0.49285
9 0.4976
10 0.5245
};
\addplot [very thin, green, mark=*, mark size=1, mark options={solid}]
table {%
1 0.91205
2 0.9002
3 0.9067
4 0.9258
5 0.89655
6 0.8698
7 0.9178
8 0.9098
9 0.87145
10 0.92205
};
\addplot [very thin, blue, dashed, mark=*, mark size=1, mark options={solid}]
table {%
1 0.593
2 0.57005
3 0.5612
4 0.62265
5 0.55715
6 0.5808
7 0.58105
8 0.5956
9 0.5522
10 0.5489
};
\addplot [ red,  densely dotted, mark=*, mark size=1, mark options={solid}]
table {%
1 0.44195
2 0.49045
3 0.3981
4 0.4637
5 0.3914
6 0.469
7 0.53575
8 0.4697
9 0.4501
10 0.51695
};
\addplot [very thin, green, mark=*, mark size=1, mark options={solid}]
table {%
1 0.8305
2 0.8774
3 0.84765
4 0.9195
5 0.91795
6 0.87605
7 0.9269
8 0.9025
9 0.81185
10 0.848
};
\addplot [very thin, blue, dashed, mark=*, mark size=1, mark options={solid}]
table {%
1 0.62715
2 0.65055
3 0.60005
4 0.65965
5 0.6858
6 0.62035
7 0.6654
8 0.68945
9 0.642
10 0.66125
};
\addplot [ red,  densely dotted, mark=*, mark size=1, mark options={solid}]
table {%
1 0.2737
2 0.3509
3 0.2682
4 0.2632
5 0.21895
6 0.2607
7 0.2802
8 0.26745
9 0.2764
10 0.256
};
\addplot [very thin, green, mark=*, mark size=1, mark options={solid}]
table {%
1 0.89885
2 0.8655
3 0.9115
4 0.7897
5 0.82915
6 0.81725
7 0.8318
8 0.8291
9 0.8122
10 0.8589
};
\addplot [very thin, blue, dashed, mark=*, mark size=1, mark options={solid}]
table {%
1 0.5302
2 0.5495
3 0.5591
4 0.56605
5 0.59015
6 0.62535
7 0.57495
8 0.5857
9 0.5712
10 0.55825
};
\addplot [ red,  densely dotted, mark=*, mark size=1, mark options={solid}]
table {%
1 0.3459
2 0.2885
3 0.22175
4 0.4151
5 0.35535
6 0.34745
7 0.32815
8 0.33415
9 0.27205
10 0.32445
};
\addplot [very thin, green, mark=*, mark size=1, mark options={solid}]
table {%
1 0.81185
2 0.77985
3 0.8581
4 0.8025
5 0.8522
6 0.857
7 0.84815
8 0.9043
9 0.9203
10 0.8302
};
\addplot [very thin, blue, dashed, mark=*, mark size=1, mark options={solid}]
table {%
1 0.56765
2 0.53645
3 0.6054
4 0.5759
5 0.64895
6 0.60405
7 0.5755
8 0.63445
9 0.5659
10 0.605
};
\addplot [ red,  densely dotted, mark=*, mark size=1, mark options={solid}]
table {%
1 0.25765
2 0.25605
3 0.2492
4 0.2419
5 0.28235
6 0.258
7 0.28605
8 0.27425
9 0.31125
10 0.2591
};
\addplot [very thin, green, mark=*, mark size=1, mark options={solid}]
table {%
1 0.76435
2 0.7776
3 0.82735
4 0.7747
5 0.86285
6 0.85
7 0.83605
8 0.86995
9 0.8293
10 0.89115
};
\addplot [very thin, blue, dashed, mark=*, mark size=1, mark options={solid}]
table {%
1 0.5049
2 0.5762
3 0.6158
4 0.583
5 0.5931
6 0.58445
7 0.5848
8 0.6589
9 0.61465
10 0.63515
};
\addplot [ red,  densely dotted, mark=*, mark size=1, mark options={solid}]
table {%
1 0.1614
2 0.1607
3 0.1484
4 0.1702
5 0.2337
6 1
7 0.26275
8 1
9 0.20085
10 0.1789
};
\addplot [very thin, green, mark=*, mark size=1, mark options={solid}]
table {%
1 0.83305
2 0.85505
3 0.85065
4 0.8802
5 0.89055
6 0.9044
7 0.85965
8 0.8549
9 0.8929
10 0.9236
};
\addplot [very thin, blue, dashed, mark=*, mark size=1, mark options={solid}]
table {%
1 0.5242
2 0.6433
3 0.60265
4 0.64135
5 0.6502
6 0.64085
7 0.62365
8 0.6182
9 0.63865
10 0.5849
};
\addplot [ red,  densely dotted, mark=*, mark size=1, mark options={solid}]
table {%
1 0.1957
2 0.2383
3 0.19145
4 0.2196
5 0.264
6 0.28715
7 0.2266
8 0.2776
9 0.2724
10 0.1933
};
\addplot [very thin, green, mark=*, mark size=1, mark options={solid}]
table {%
1 0.9512
2 0.9707
3 0.95715
4 0.9611
5 0.96495
6 0.96205
7 0.9704
8 0.96975
9 0.93445
10 0.96535
};
\addplot [very thin, blue, dashed, mark=*, mark size=1, mark options={solid}]
table {%
1 0.51785
2 0.5354
3 0.54985
4 0.5217
5 0.5671
6 0.5417
7 0.5259
8 0.55605
9 0.56485
10 0.56245
};
\addplot [ red,  densely dotted, mark=*, mark size=1, mark options={solid}]
table {%
1 0.29085
2 0.31515
3 0.27275
4 0.36365
5 0.37225
6 0.37015
7 0.3599
8 0.36105
9 0.4233
10 0.361
};
\end{axis}

\end{tikzpicture}
\caption{The malware Recall rate for the Easy (green), Medium (blue), and Hard (red) splits for MalConvGCT as training progresses. Note the two spikes are failure cases of the model marking all files as ``malware'' (100\% false-positive rate). The results show how our approach produces multiple splits that are in a grouped range of similar difficulty. }
\label{fig:MalConvGCTCombinedSplitsBenignAccuracyFixed0.8}
\end{figure}

Recall that the same splits are being used for all four algorithms. This shows an unusual, but important, kind of generalization. Even though MalConv is less accurate in normal use than MalConvGCT, and significantly less accurate than domain-knowledge-wielding XGBoost, the Hard split is able to reduce XGBoost down to just 72.80\% accuracy. This shows that our benchmark search is 1) finding correlations of intrinsic difficulty, and 2) allowing us to avoid overly biasing a test-set against a specific approach. That is to say, if we used an XGBoost model to produce the splits to evaluate an improved XGBoost, we may unfairly over-compensate by having produced a dataset split that is too difficult. 

To show the consistency of our results in producing train/test splits of comparable difficult, we show the result for multiple splits as a function of how many epochs MalConvGCT has trained for in \autoref{fig:MalConvGCTCombinedSplitsBenignAccuracyFixed0.8}. Here it is clear that each of the Easy, Medium, and Hard difficulty levels exhibit high degrees of similarity in their difficulty. This is important to avoid a naive solution where the target difficulty is obtained by averaging models that are \textit{too} hard against others that are \textit{too} easy. Such an undesirable scenario would make performing multiple trials to use statistical tests difficult, as the overly easy and hard splits would degrade to adding noisy samples\footnote{Because each model easily gets all the easy splits correct, and the hard splits all misclassified, making the differences between two models indistinguishable.} to the test and reduce the total power of the test to conclude if one method was really better than another~\cite{JMLR:v17:benavoli16a}. 

\subsection{Ablation} \label{sec:ablation}

Having established the efficacy of our approach to producing datasets of the desired difficulty level, we will now demonstrate two alternative but intuitive strategies that do not meet our needs. In the case of a desired ``Easy'' benchmark, one may naively select the top-$K$ ``best'' families from \autoref{fig:cross_errors}, which have the highest average recall against other malware families. Second, one may similarly decide that a ``Hard'' dataset should be produced by selecting the families with the lowest average recall.

\begin{figure}[!ht]
\begin{tikzpicture}

\definecolor{chocolate2267451}{RGB}{226,74,51}
\definecolor{dimgray85}{RGB}{85,85,85}
\definecolor{gainsboro229}{RGB}{229,229,229}
\definecolor{steelblue52138189}{RGB}{52,138,189}

\begin{axis}[
axis background/.style={fill=gainsboro229},
axis line style={white},
tick align=outside,
tick pos=left,
width=10cm,
x grid style={white},
xlabel=\textcolor{dimgray85}{Top K Families Model},
xmajorgrids,
xmin=-0.3, xmax=6.3,
xtick style={color=dimgray85},
xtick={0,1,2,3,4,5,6},
xtick={0,1,2,3,4,5,6},
xticklabels={5,10,15,20, 25, 30, 35},
y grid style={white},
ylabel=\textcolor{dimgray85}{Average Recall value},
ymajorgrids,
ymin=0.70, ymax=0.85,
ytick style={color=dimgray85},
height=0.5\columnwidth,
width=\columnwidth,
]
\addplot [line width=0.44pt, chocolate2267451, mark=*, mark size=3, mark options={solid}]
table {%
0 0.713698369565217
1 0.785750000000001
2 0.786861413043479
3 0.722097826086956
4 0.711790760869565
5 0.797453804347826
6 0.810891304347826
};
\addplot [line width=0.44pt, steelblue52138189, mark=*, mark size=3, mark options={solid}]
table {%
0 0.713698369565217
1 0.785750000000001
2 0.786861413043479
3 0.722097826086956
4 0.711790760869565
5 0.797453804347826
6 0.810891304347826
};
\end{axis}

\end{tikzpicture}
\caption{Average recall performance (y-axis) of MalConv Model trained on best k generalizing families (x-axis) across all 184 malware families. The performance of the model fluctuates between 71\% and 81\% in a non-monotonic fashion, making top-$K$ selection unreliable to a specific level of performance, and with little total average variation. This fluctuation makes this strategy ineffective for building a benchmark of a desired level of difficulty. }
\label{fig:averagebestplot}
\end{figure}
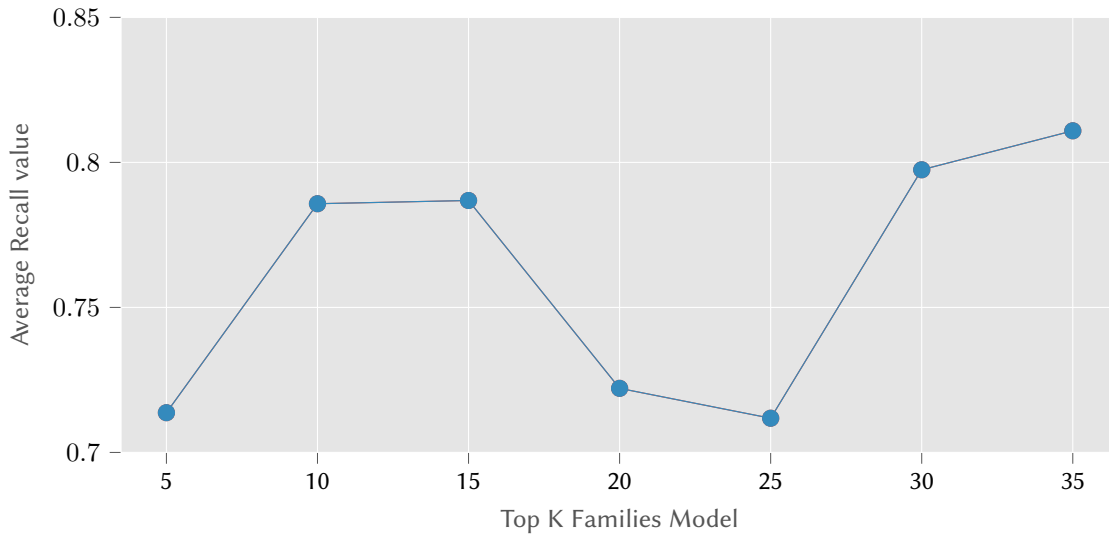

For the ``Easy'' case of selecting the top-$K$, we first show as an example the results of this strategy when picking the top-5 most frequent families in \autoref{fig:best5}. Though this produces a recall of 70\%, the variance of the results is extremely high. This huge variance is undesirable for the same reason as our results from \autoref{fig:MalConvGCTCombinedSplitsBenignAccuracyFixed0.8}.  
We want reasonably similar performance characteristics for each split to maximize the power of subsequent conclusions about improvement. Each overly easy and hard split is one that does not provide meaningful information to the question of whether a new algorithm would perform better. 

One may wonder instead if the issue would improve by selecting more families. This is unfortunately not the case, and there is relatively little variation as the top-$K$ is altered from $K=5$ to $K=35$, as shown in \autoref{fig:averagebestplot} (We note that all values of $K$ look qualitatively similar to \autoref{fig:best5} as well).

\begin{figure}[!ht]
\begin{tikzpicture}

\definecolor{dimgray85}{RGB}{85,85,85}
\definecolor{gainsboro229}{RGB}{229,229,229}
\definecolor{steelblue51137188}{RGB}{51,137,188}

\begin{axis}[
axis background/.style={fill=gainsboro229},
axis line style={white},
tick align=outside,
tick pos=left,
width=10cm,
x grid style={white},
xlabel=\textcolor{dimgray85}{Malware Families used for Testing},
xmajorgrids,
xmin=0, xmax=183,
xtick style={color=dimgray85},
y grid style={white},
ylabel=\textcolor{dimgray85}{Recall value},
ymajorgrids,
ymin=0.0, ymax=1.05,
ytick style={color=dimgray85},
height=0.5\columnwidth,
width=\columnwidth,
]
\addplot [line width=0.44pt, steelblue51137188]
table {%
0 0.998
1 0
2 0.0005
3 0
4 0
5 0
6 0
7 0
8 0
9 1
10 0
11 0
12 0
13 0
14 0
15 0.0005
16 0
17 0
18 0
19 0
20 0
21 0.995
22 0
23 0
24 0.0005
25 0
26 0
27 0
28 0
29 0
30 0
31 0
32 0
33 0
34 0
35 0
36 0
37 0
38 0
39 0
40 0
41 0
42 0
43 0
44 0
45 0
46 0
47 0
48 0
49 0.9995
50 0
51 0
52 0
53 0
54 0
55 0
56 0
57 0
58 0
59 0
60 0
61 0
62 0
63 0
64 0
65 0
66 0
67 0
68 0
69 0.997
70 0.004
71 0
72 0.0005
73 0.003
74 0
75 0
76 0
77 0
78 0
79 0
80 0
81 0
82 0
83 0
84 0
85 0
86 0
87 0
88 0.993
89 0
90 0
91 0
92 0
93 0.003
94 0
95 0
96 0
97 0
98 0
99 0
100 0
101 0
102 0
103 0
104 0
105 0.0055
106 0
107 0
108 0
109 0
110 0
111 0
112 0.0005
113 0
114 0
115 0.0005
116 0
117 0
118 0
119 1
120 0
121 0
122 0
123 0
124 0.0005
125 0
126 0
127 0
128 0
129 0
130 0
131 0
132 0
133 0
134 0
135 0
136 0
137 0
138 0
139 0
140 0
141 0
142 0
143 0
144 0
145 0.0005
146 0
147 0
148 0
149 0
150 0.002
151 0
152 0
153 0
154 0
155 1
156 0
157 0
158 1
159 0
160 0
161 0
162 0
163 0
164 0
165 0
166 0
167 0
168 0
169 1
170 0
171 0
172 0
173 0
174 0
175 0
176 0
177 0
178 0
179 0
180 0.0005
181 0
182 0
183 0
};
\addplot [line width=0.44pt, steelblue51137188, mark=*, mark size=1, mark options={solid}, only marks]
table {%
0 0.998
9 1
21 0.995
49 0.9995
69 0.997
88 0.993
119 1
155 1
158 1
169 1
};
\end{axis}

\end{tikzpicture}
\caption{Malware family (x-axis, by ID number for space) and the recall of that family (y-axis) of a MalConv ~\cite{MalConv} Model trained on worst 10 families. The result has almost zero generalization to any other malware family, and the 10 high recalls of near 100\% correspond to the 10 families used to train the model. This shows selecting the worst generalization families is too hard to construct a useful benchmark.}
\label{fig:worst10plotfamily}
\end{figure}
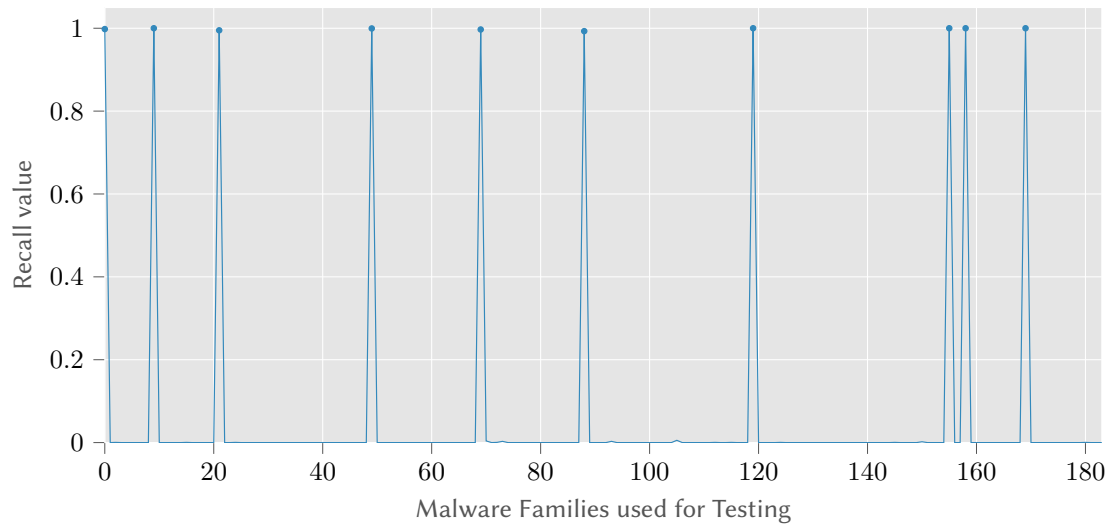

A different kind of issue occurs when selecting the worst-$K$ malware families to produce a ``Hard'' dataset. $K=10$ is shown as an example in \autoref{fig:worst10plotfamily}, where the 10 chosen families each have 100\% recall, and the model does not meaningfully learn to detect any of the remaining malware families. In this case, the hardest families are so distinct on their own that the model easily learns to overfit to the specific malware families, and the default for any other input 
becomes ``benign''. This is similar to the overly-strong data leakage signal discussed by \cite{raff_ngram_2016} when building a benign dataset from scraping a clean install of Microsoft Word. We again note that using multiple values of $K$ all result in qualitatively the same results for the worst-$K$ strategy. 

\section{Conclusion} \label{sec:conclusion}

We have now shown it is possible to use malware family information to construct better train/test splits for benchmarking purposes, where the difficulty of the split is configurable. This was demonstrated with an Easy, Medium, and Hard split --- and in all cases a weaker model is able to produce splits that are effective against a more powerful model. This is a necessary condition of utility, as the purpose of the splits is to test a hopefully more powerful alternative model. We further validate our approach by ablating against simpler design alternatives, which do not produce benchmarks of usable quality.

\bibliography{figures/references}

\end{document}